# *KOM: A Multi-Agent Artificial Intelligence System for Precision Management of Knee Osteoarthritis (KOA)*

Weizhi Liu[1#]; Xi Chen[2#]; Zekun Jiang[3,4,5#]; Liang Zhao[6]; Kunyuan Jiang[3,4,5]; Ruisi Tang[1]; Li Wang[1]; Mingke You[1]; Hanyu Zhou[7]; Hongyu Chen[8]; Qianjiang Xiong[2]; Yong Nie[9*]; Kang Li[3,4,5*]; Jian Li[2*]

[1] West China School of Medicine, Sichuan University, Chengdu, Sichuan, China

[2] Sports Medicine Center, Department of Orthopedics and Orthopedic Research Institute, West China Hospital, Sichuan University, Chengdu, Sichuan, China

[3] West China Biomedical Big Data Center, West China Hospital, Sichuan University, Chengdu, Sichuan 610041, China

[4] Sichuan University Pittsburgh Institute, Chengdu, Sichuan, China

[5] Shanghai Artificial Intelligence Laboratory, Shanghai, China

[6] Dyania Health, Boston, Massachusetts, United States of America

[7] School of Computer Science, Carnegie Mellon University, Pittsburgh, Pennsylvania, United States of America

[8] Faculty of Science, Universiteit van Amsterdam, Amsterdam, Netherlands

[9] Department of Orthopedic Surgery and Orthopedic Research Institute, West China Hospital, Sichuan University, Chengdu, Sichuan, China

#These authors contributed equally

*Co-Corresponding Authors.

Professor Yong Nie

Department of Orthopedic Surgery and Orthopedic Research Institute, West China Hospital, Sichuan University, Chengdu, Sichuan, China

Email: nieyong1983@wchscu.cn

Professor Kang Li

West China Biomedical Big Data Center, Med-X Center for Informatics, Sichuan University, Chengdu, Sichuan, China

Sichuan University Pittsburgh Institute, Chengdu, Sichuan, China

Shanghai Artificial Intelligence Laboratory, Shanghai, China

Email: likang@wchscu.cn

Telephone: 8619983138590

Professor. Jian Li,

Sports Medicine Center, West China Hospital, West China School of Medicine, Sichuan University, Chengdu, Sichuan, China.

Department of Orthopedics and Orthopedic Research Institute, West China Hospital, Sichuan University, Chengdu, Sichuan, China.

Email: lijian_sportsmed@163.com

Telephone: 8618980601388

# Abstract

Knee osteoarthritis (KOA) affects over 600 million people worldwide, causing substantial pain, functional limitations, and disability. Although tailored multidisciplinary interventions can slow disease progression and improve quality of life, they often demand considerable medical resources and expertise, limiting their feasibility in resource-constrained settings. To bridge this gap, we developed KOM (Knee Osteoarthritis Manager), a multi-agent system that automates KOA evaluation, risk prediction, and treatment prescription, enabling clinicians to perform key tasks across the KOA care continuum and generate management plans based on patient characteristics, disease status, risk factors, and contraindications. In benchmarking experiments, KOM outperformed several general-purpose large language models in imaging analysis and prescription generation. A randomized three-arm simulation study further showed that KOM–clinician collaboration reduced overall diagnostic and planning time by 38.5% and yielded higher treatment quality than either approach alone. These results suggest that KOM may help support automated KOA management and, when integrated with clinical practice, could potentially improve care efficiency. The modular design may also be informative for the development of AI-assisted management systems targeting other chronic diseases.

## Introduction

KOA affects approximately 600 million people worldwide, characterized by progressive pain and functional deterioration that frequently necessitates total knee replacement in the end [1-3]. Timely and appropriate intervention is essential for slowing structural deterioration, alleviating symptoms, and improving functional outcomes[4]. However, delivering multidisciplinary personalized KOA management for large patient groups remains challenging, particularly in resource-limited healthcare systems[3].

Advances in AI have created new opportunities for KOA management through automated analysis, risk prediction[5-11]. AI-driven radiographic techniques, particularly convolutional neural networks, now facilitate the automated analysis of osteoarthritis images[5-7]. In parallel, prediction models integrating clinical and imaging data have been developed to estimate the risk of structural progression and functional decline in KOA patients[8-10]. However, these studies have not directly demonstrated their impact on the complete KOA management workflow.

Moreover, the principal challenges in complete KOA management stem from the intensive nature of patient interactions and treatment planning. First, completing patient assessment process requires substantial clinical resources. In clinical workflows, multiple rounds of communication between clinicians and patients are typically needed to complete a full medical history collection. This process consumes significant time from clinicians, who are already facing heavy workloads, and the repeated execution of these time-consuming procedures may lead to the omission of critical information during interactions, potentially resulting in serious medical events. Second, effective disease management strategies require the formulation of personalized intervention plans based on the current disease status, quantified key progression risk factors, contraindications, and patient-specific needs. In the absence of AI assistance, it is often difficult to quickly determine key risk factors when multiple risk factors are present, and it is also challenging to formulate an individualized management plan for knee osteoarthritis based on the patient's information within a short period.

To address these limitations, we have explored several solutions. Our early work developed structured prompting techniques to enhance LLM performance for osteoarthritis queries[12]. An osteoarthritis agent (DocOA) used LLMs with Retrieval-Augmented Generation (RAG) to access guideline knowledge and generate compliant treatment recommendations[13]. After that, a multi-agent system was developed for complex clinical cases, where agents simulate different specialists in multidisciplinary team discussions[14]. These incremental developments provided preliminary insights but demonstrated limited gains in overall KOA care quality and efficiency, still failing to alleviate the clinical burden of KOA management. However, these findings motivated the production of the KOM system, a multi-agent AI system designed to support multiple components of KOA management, including patient interaction and assessment, risk prediction, and individualized treatment planning. KOM is implemented using a modular and extensible architecture designed to support future adaptation to other chronic diseases requiring complex, longitudinal management. The system consists of three specialized agents integrating LLMs, ResNet architecture, and other machine learning algorithms. Specifically, it comprises:

1. An Assessment Agent capable of interacting with patients, processing multimodal data, analyzing radiological images, and generating structured evaluation reports.

2. A Risk Agent designed to extract patient-specific progression risk factors, predicts individualized KOA progression, and generates risk reports.

3. A Therapy Agents Group consists of a set of domain-specific agents maintaining evidence-based medical knowledge, collectively simulating multidisciplinary team (MDT) discussions to generate personalized management plans.

Following the clinical workflow, KOM completes the process from data collection to management planning. Specifically, the Assessment Agent collects and evaluates a patient's demographic information, present and past medical history, personal history, imaging data, psychological state, nutritional status, physical activity level, socioeconomic condition, and treatment preferences, and automatically generates a structured electronic medical report. Subsequently, the Risk Agent extracts relevant clinical indicators to predict short- and mid-term structural and symptomatic progression of KOA, and identifies quantified risk factors that enable data-driven risk stratification and disease monitoring. Based on the medical reports and risks reports from previous two agents, the Therapy Agents Group further simulates an MDT decision-making process to generate intervention plans. Each agent was evaluated against general-purpose language models and algorithmic baselines to assess performance.

Furthermore, we conducted a three-arm comparative study to assess the system's clinical utility in controlled simulated environments. The comparative study included three groups: an independent KOM deployment (KOM alone), an integrated KOM-doctoral collaboration (KOM plus clinicians), and a traditional doctoral practice (clinicians alone).

This study describes the development, validation, and clinical evaluation of KOM and examines its potential role in enhancing the quality and efficiency of KOA management.

## Results

**Development of the Knee Osteoarthritis Management (KOM) System.**

KOM is an interactive multi-agent system that supports clinicians in three areas for KOA management: patient information collection and assessment, disease trajectory prediction and etiology analysis, and individualized treatment planning (Figure 1).

The KOM architecture features three specialized agents:

1.Assessment Agent: Features multi-round patient interaction capabilities, image analysis functionality, and summary report generation. collects and evaluates a patient's demographic information, present and past medical history, personal history, imaging data, psychological state, nutritional status, physical activity level, socioeconomic condition, and treatment preferences. It analyzes knee radiographs to classify the severity of KOA and identify specific features, including osteophyte formation and alterations in joint space. Finally, The Assessment Agent generates a structured evaluation report (Figure 2a).

2.Risk Agent: Capable of predicting knee osteoarthritis progression and identifying individual risk factors that contribute to disease advancement. It extracts parameters from evaluation report to forecast KOOS (Knee Injury and Osteoarthritis Outcome Score) subscale scores and KL (Kellgren–Lawrence) radiographic grading in the following four years. Finally, the Risk Agent generates a structured risk report (Figure 3a).

3.Therapy Agents Group: Composed of specialist agents from different medical disciplines, each with specialized knowledge bases. These agents simulate clinical physicians in multidisciplinary team discussions and develop multidisciplinary treatment plans based on evaluation report and risk report (Figure 4a).

The workflow of KOM is illustrated in Figure 1b. The system offers two interaction modalities: sequential progression through the complete pathway or independent access to individual agents with manual data input. This flexibility accommodates diverse clinical workflows across healthcare settings.

**Assessment Agent**

The Assessment Agent collects patient information through structured dialogues and performs radiographic image analysis to generate evaluation reports. Its workflow is organized into three stages: information collection, treatment willingness confirmation, and radiographic analysis (Figure 2a). The agent integrates two key components: a patient–agent interaction module and an X-ray analysis module. The radiographic analysis output is seamlessly incorporated into the patient history, enabling a unified and structured clinical record.

For the Patient-agent interaction module, this study implemented Qwen-Max as the foundation model. It developed a structured prompt through systematic engineering to facilitate multi-turn conversations between the agent and patients. This approach enabled the collection of information and the generation of summary reports. The model was configured with a temperature parameter of 0.8 and accessed via the application programming interface (API). Three human physicians evaluated 100 simulated patient-agent interactions, in which physicians acted as patients interacting with the assessment agent. The human physicians evaluated the interaction process and the final summary report, rating them on four metrics on a scale of 1 to 5, where 1 indicated no compliance and 5 indicated complete compliance. The results are as follows (Figure 2g): field completeness (4.03 ± 0.21), logical consistency (3.99 ± 0.16), medical accuracy (4.37 ± 0.47), and readability (4.00 ± 0.23).

The X-ray analysis module was developed using the Osteoarthritis Initiative (OAI) dataset, which contains longitudinal bilateral knee radiographs from 4796 cases at their baseline evaluation, 2-year follow-up evaluation, and 4-year follow-up evaluation, resulting in a total of 12,719 bilateral anterior-posterior knee X-rays. The X-ray analysis module performs KOA severity grading, detects bilateral osteophytes, and assesses bilateral joint space. The workflow consists of initial knee center localization to determine the region of interest (ROI), followed by identification of both the left and right knee joints, with subsequent osteophyte detection and joint space analysis for each joint. To implement these functionalities, we developed a series of algorithms trained on a randomly selected subset of data from the OAI dataset, which human experts had calibrated before model training. The knee center localization algorithm utilized a U-Net architecture trained on 200 labeled images. After 40 training epochs, the validation metrics improved: the loss decreased from 0.7576 to 0.3804, the IoU increased from 0.0016 to 0.5790, and the center point error was reduced from 146.02 to 4.83 pixels (Figure 2d). For severity classification, a ResNet model was trained on balanced class distributions. Using an 8:1:1 data partition with 5-fold cross-validation, the model achieved an overall accuracy of 80.8% (Figure 2e). The None/Doubt class showed the highest accuracy (90.7%), with confusion primarily between Moderate and Mild grades. We also developed ten specialized models for extracting radiographic features from distinct anatomical regions, including the medial and lateral joint spaces, as well as the medial and lateral aspects of both femoral and tibial surfaces. The lateral joint space narrowing classification model achieved an accuracy of 89.8%, surpassing the medial joint space narrowing model (77.1%). Classification accuracy for subchondral sclerosis demonstrated regional variation, with the highest accuracy observed in the medial tibial plateau (56.4%), the lateral tibial plateau (83.1%), the medial femoral condyle (60.6%), and the lateral femoral condyle (85.3%). Osteophyte detection models maintained relatively consistent performance across all anatomical quadrants, with accuracy ranging from 78.5% to 95.5%. Gradient-weighted Class Activation Mapping (Grad-CAM) analysis enhanced model interpretability for classification tasks (Figure 2f). The relevant model design, training processes, and detailed the Supplementary Graphs S2.

**Comparative Evaluation of Radiographic Performance**

The Assessment Agent was benchmarked against five leading vision-language models (Google Gemini 2.0 Pro, GPT-4o, Claude 3.7, QwenMax VL, and LLaMA 3.2 90B Vision Instruct) using 500 bilateral knee radiographs from the OAI dataset, which were excluded from the training set. The evaluation assessed KOA severity grading, the detection of OA presence. For KOA severity grading (Figure 2b, c, h), the Assessment Agent achieved 77.16% accuracy, outperforming Gemini 2.0 Pro (34.50%). In OA presence detection, the Assessment Agent attained an accuracy of 82.22% compared to Gemini's 76.66%. The Assessment Agent maintained performance across anatomical locations with 75.38% and 78.95% accuracy on left and right KOA severity grading, respectively, and 84.09% and 80.35% accuracy for left and right knee OA detection. All competing models showed lower accuracy below 65% on these tasks.

Moreover, the Assessment Agent demonstrated diagnostic capability across different levels of disease severity (64.68%-82.16% accuracy across classifications). In contrast, competing models often achieve high accuracy for None/Doubt cases but poor performance (<20%) on Mild or higher grades (Figure 2c), which may constrain their applicability in similar clinical tasks. Detailed metrics for this task are available in the Supplementary Graphs S2.

**Risk Agent**

The Risk Agent predicts the functional outcome and radiographic outcome of KOA at 1 years and 4 years of follow-up. It also identifies patient-specific risk factors.

At the 1-year follow-up (V01), all six machine learning models demonstrated predictive ability across KOOS subscores (Figure 3d). The strongest results were obtained for right knee symptoms and quality of life, with correlation coefficients approaching 0.74 and explained variance ($R^2$) values exceeding 0.50 in the best-performing models. ElasticNet provided relatively stable performance across multiple KOOS subscores, achieving $R^2$ values up to 0.58 with relatively low mean absolute errors. Random Forest and Gradient Boosting also performed well, particularly in predicting pain and function-related scores. In contrast, SVR and LightGBM showed less consistent results. At the 4-year follow-up (V06), predictive accuracy declined across all KOOS subscores. The best-performing models reached correlation values of 0.65–0.69 with R² values between 0.30 and 0.46, notably lower than at V01. Random Forest, Gradient Boosting, and ElasticNet remained relatively stable performance across tasks, while SVR and LightGBM again produced weaker predictions. Despite the decline, the prediction of quality-of-life and pain subscores retained moderate correlation values, whereas sports and recreation scores showed the lowest stability.

For KL Grade Classification prediction tasks (Figure 3b, c) eight algorithms were evaluated. At V01, ensemble-based models achieved the highest predictive performance. For the left knee, AdaBoost reached the best overall results (accuracy = 0.910, F1 = 0.908, AUC = 0.965). Random Forest (accuracy = 0.902, AUC = 0.971) and XGBoost (accuracy = 0.897, AUC = 0.965) also performed strongly. For the right knee, LightGBM (accuracy = 0.910, AUC = 0.962) and XGBoost (accuracy = 0.908, AUC = 0.967) achieved the highest classification accuracy, while Random Forest remained highly competitive (accuracy = 0.900, AUC = 0.972). Across both knees, all ensemble approaches produced AUC values above 0.96, indicating robust discriminative capacity. At V06, predictive performance declined compared with V01. For the left knee, Gradient Boosting, XGBoost, and LightGBM produced the most consistent results, with accuracies ranging from 0.75 to 0.76 and AUC values close to 0.92. For the right knee, XGBoost yielded the best balance of metrics (accuracy = 0.765, AUC = 0.922), while Random Forest also performed well (accuracy = 0.762, AUC = 0.932). Other algorithms, including SVM, neural networks, and KNN, exhibited weaker performance at both time points.

Following functional outcome prediction, individualized risk factors were identified using SHAP analysis, which enhanced interpretability by quantifying the contributions of each feature to the prediction. In a representative case, the predicted KOOS symptom score (72.18) fell below the cohort mean (75.50), with primary negative contributors including osteophytes in baseline left knee X-ray (-1.08), diminished KOOS pain score (-1.02), suggesting osteophytes and pain are the individualized progression risks for this patient. This patient also shows below-average peak knee extension torque; this parameter contributed positively (+1.19), suggesting residual muscle strength may protect against symptomatic progression. A detailed description of the model design, training procedures, and experimental results is provided in Supplementary Graphs S3 as well as Supplementary Tables T1 and T2.

**Therapy Agents Group**

A multi-agent cluster was developed to facilitate multi-agent conversations, mimicking the Multi-Disciplinary Team (MDT) approach adopted in clinical practice, for formulating patient-specific, multidisciplinary management plans. The cluster of agents receives the evaluation report and risk report generated in previous stages, engages in discussion, and generates the final personalized management plan. The cluster comprised multiple agents functioning as specialists, including an Exercise Rehabilitation agent, an Orthopedic agent, a Psycho-Nutrition agent, and a Clinical Decision agent.

The Clinical Decision Agent functions as the coordinator and integrator within the cluster. It synthesizes the recommendations provided by the other specialist agents, resolves conflicts where their suggestions diverge, and applies evidence-based clinical guidelines to ensure the final plan is coherent, feasible, and clinically appropriate. In addition, the Clinical Decision Agent prioritizes interventions based on patient risk factor, comorbidities, and treatment preferences, thereby generating a plan that aligns with real-world clinical decision-making standards. Each agent was furnished with domain-specific medical data. Qwen-Max was utilized as the base model for all agents, with a temperature parameter set at 0.8. The model was accessed via API. Prompt engineering was conducted to instruct each agent to act as a clinical specialist, engage in active discussion with other agents, and develop a personalized management plan for the given KOA patient. Each agent is equipped with a retrieval augmented generation tool to utilize medical data from their respective knowledge base to generate and revise the management plan. We curated six specialized medical databases; each derived from authoritative clinical guidelines and peer-reviewed articles indexed in the Medline database (Figure 4a). All databases were constructed through a structured pipeline of literature search, eligibility screening, data extraction, and knowledge structuring. Each agent within the multi-agent cluster is paired with its corresponding evidence database to generate patient-specific recommendations:

KOM Agent 1 – Nutrition and Psychology: linked to the psychological database (210 entries) and nutrition database (349 entries), enabling the generation of individualized psychological counseling and nutritional prescriptions.

KOM Agent 2 – Medication and Surgery: connected to the surgical evidence database (1,549 entries) to determine surgical indications and medication strategies based on established osteoarthritis guidelines.

KOM Agent 3 – Exercise Prescription: supported by the rehabilitation database (934 entries) and Exercise database (975 entries), which provides evidence-based exercise regimens tailored to the patient's KOA severity and physical capacity.

KOM Agent 4 – Clinical Decision and Summary: serves as the coordinator and synthesizer, using a shared guideline database to integrate the recommendations of all other agents, resolve conflicts, and formulate a coherent, evidence-based management plan.

Additionally, a guideline database is accessible to all agents, ensuring that each recommendation aligns with up-to-date clinical practice standards.

We evaluated the quality of generated treatment plans using retrospective clinical data from 250 patients with knee osteoarthritis treated at West China Hospital. For each case, we conducted expert evaluations of the generated prescriptions and calculated their similarity scores against gold-standard prescriptions to assess system performance. And we benchmarked KOM against leading general-purpose AI models, including GPT-4o, GPT-4o-mini, DeepSeekR1, Claude 3.7 Sonnet, QwenMax, Qwen2.5-14B, and Gemini 2.0 Pro (Figure 4a).    Additionally, we benchmarked a single-agent RAG implementation against our agents' group with domain-specific databases and collaborative decision-making.

For lexical and semantic similarity analysis, we employed three established metrics. KOM achieved the highest BLEU score (0.0191), outperforming GPT-4o (0.0064) and Qwen2.5-14B (0.0083). Similarly, KOM led in ROUGE-L metrics with 0.2905, higher than QwenMax (0.1244) and DeepSeekR1 (0.1031). BERT evaluations showed narrower differences, with KOM (0.8069) performing comparably to GPT-4o-mini (0.8156) and GPT-4o (0.8122), demonstrating competitive semantic consistency across models (Figure 4g).

The expert evaluation involved three specialists in orthopedics and sports medicine who independently rated prescriptions across seven dimensions on a 1–5 scale. KOM achieved the highest composite score (29.63 ± 1.33), outperforming the next-best model, DeepSeekR1 (26.03), by 3.60 points (Figure 4b, c). KOM received higher mean ratings across all evaluated dimensions, including completeness (4.408), personalization (4.380), and safety (4.366). Although nutritional guidance represented the lowest-scoring dimension across all models, KOM maintained its leading position with a score of 3.903 (Figure 4d, e).

Z-score normalization (Figure 4f) highlighted KOM's relative strengths in completeness (+0.67), personalization (+0.73), and safety (+0.66), with moderate performance in evidence-based practice (+0.48) and feasibility (+0.06). Other models demonstrated specific advantages in individual domains: G4M in evidence-based practice (+0.99), GPT-4o-mini in completeness (+0.76) and safety (+0.87), QwenMax in safety (+1.13), DeepSeekR1 in personalization (+1.10), and Gemini 2.0 Pro in completeness (+1.10). Across all models, exercise design and nutritional advice consistently emerged as weaker areas of focus. The relevant model design, training processes, and experimental result data are presented in detail in Figure 4 and the Supplementary Graphs S4 for reference.

**Clinical Evaluation of the KOM System**

To evaluate the effectiveness of the KOM system in clinical practice, we conducted an end-to-end simulation using 50 cases of KOA from West China Hospital (Figure 5a). The clinical evaluation included three operating conditions: physicians performing the assessment and treatment planning process alone, the KOM system functioning autonomously and performing the process, and physicians collaborating with the KOM system, where the system performs the X-ray evaluation and treatment planning, and the physicians can supervise and modify the reports generated by the KOM system at each stage of the process (Figure 5b). The quality of X-ray assessment, the quality of the management plan, and the entire processing time were evaluated.

The approval rate of radio-graphic grading was defined as the proportion of knee images correctly classified according to severity grading within the final cohort of 50 cases. Expert evaluation of KOA grading results showed that image classifications generated independently by ten physicians achieved approval rates ranging from 42.0% to 66.0%, with a mean approval rate of 56.0%. When assisted by the KOM system, the approval rates increased, ranging from 90.0% to 96.0%, yielding a mean approval rate of 93.0%. Under the fully automated KOM-only condition, approval rates ranged from 72.0% to 82.0%, corresponding to a mean approval rate of 77.4% (Figure 5e). Expert evaluation was conducted on the same 50 prescriptions, with each prescription assessed across seven clinical criteria: clinical evidence, completeness, exercise prescription standardization, nutritional prescription standardization, safety, personalization, and accessibility. The aggregate average score was 3.63 for MS (Physicians), 4.56 for KOM, and 4.43 for the collaboration group (Figure 5d). Regarding content completeness, the MS+KOM group achieved a score of 4.73, compared with 4.63 for KOM and 4.01 for MS. For personalization, both the collaboration group and KOM group demonstrated comparable performance, with scores exceeding 4.80. Similarly, for safety, both groups maintained scores above 4.80. In exercise prescription quality, the scores were 3.13 (MS), 4.11 (KOM), and 4.10 (MS+KOM), highlighting the benefit of AI augmentation. For nutritional guidance, the scores were 3.30 (MS), 3.93 (KOM), and 3.97 (MS+KOM). In accessibility and feasibility, the collaboration group scored 4.10, lower than KOM's 4.59. In adherence to evidence-based practice, the collaboration group achieved 4.41, while KOM scored 4.63.

Quantitative prescription similarity metrics demonstrated that BLEU scores were 0.0065 (MS), 0.0455 (KOM), and 0.0500 (collaboration group). ROUGE-L scores were 0.1126 (MS), 0.2590 (KOM), and 0.2340 (collaboration group). BERTs were 0.8021 (MS), 0.7996 (KOM), and 0.8116 (collaboration group), indicating superior semantic alignment with reference standards in the human-AI collaborative condition (Figure 5f).

For the complete clinical workflow, the MS group required $586 \pm 56$ seconds per case. In contrast, the collaboration group completed identical tasks in $361 \pm 42$ seconds (Figure 5c), demonstrating a 38.5% reduction in processing time.

# Discussion

## Main Finding

This study introduces the KOM system; the first evaluated multi-agent systems for KOA. The Assessment Agent acquires patient-related information and treatment goals through interactive dialogue and analyzes knee radiographs to classify KOA severity in accordance with established clinical criteria. The Risk Agent forecasts functional and radiographic outcomes at one- and four-years follow-up while identifying patient-specific risk factors to inform intervention planning. The Therapy Agents Group, designed to simulate multidisciplinary team discussions, generates evidence-based, personalized management plans by integrating domain-specific knowledge from rehabilitation, exercise, surgical, psychological, and nutritional specialties. Evaluation results indicate that the Assessment Agent demonstrates enhanced performance compared to general-purpose AI models across assessment parameters. The Risk Agent accurately predicted functional and radiographic outcomes at 1-year and 4-year follow-ups. The Therapy Agents Group developed evidence-based, individualized management plans that demonstrated higher quality than those generated by current language models across multiple evaluation metrics.

In a clinical evaluation study comprising three groups (physicians alone, KOM alone, and doctoral trainee-KOM collaboration), the collaboration group demonstrated significant advantages. Expert reviewers approved 93.0% of treatment plans from the doctoral trainee-KOM collaboration compared to 53.8% for physicians alone and 77.4% for KOM alone. Quality assessment across seven clinical criteria revealed superior performance in the collaborative condition, particularly in content completeness, personalization, and safety considerations. Notably, the doctoral trainee-KOM collaboration resulted in a 38.5% reduction in processing time compared to physicians working independently.

These findings suggest that KOM may serve as a useful clinical decision support tool that can enhance both the quality and efficiency of KOA management while providing a methodological framework for developing similar systems for other chronic degenerative disorders.

**Assessment Agent via LLM-DL Hybrid Architecture**

The Assessment Agent of the KOM system employs a hybrid architecture that integrates deep learning-based image interpretation with prompt-optimized LLM. The system utilizes a ResNet-based convolutional neural network trained on more than 12,000 standardized bilateral anteroposterior knee radiographs from the OAI database. This enables classification of KOA severity, alongside accurate assessments of medial and lateral joint space narrowing, osteophyte presence, and subchondral bone sclerosis.

While general-purpose vision-language models (VLMs) such as GPT-4V demonstrate zero-shot generalization capabilities in open-domain tasks, our findings reveal significant limitations when applied to specialized clinical imaging interpretation[15-18]. In our evaluation, these models demonstrated inadequate accuracy and consistency in grading KOA severity under zero-shot conditions. Traditional machine learning techniques, including gradient-boosted trees and convolutional neural networks trained on curated OA datasets, have repeatedly shown performance comparable to expert readers in radiographic grading.[19-21]. For patient information collection, LLMs have demonstrated effectiveness in generating structured clinical narratives and supporting interactive clinical workflows[22]. Recognizing these strengths, we developed a hybrid LLM-DL framework that produces a flexible, interpretable, and clinically applicable workflow for case generation. This approach aligns with recent developments in hybrid architectures, such as DeepDR-LLM, which combined image-based transformers with LLMs fine-tuned on 370,000 real-world diabetes management records to generate personalized recommendations[23]. Those suggest that LLM-DL architectures provide an adaptable and controllable solution for structured documentation in medical domains with defined task parameters and standardized inputs.

**Risk Agent with Etiological Analysis**

The Risk Agent in KOM forecasts both symptomatic and structural trajectories of KOA using supervised machine learning algorithms. This component models the temporal changes in KOOS subdomains and KL grades at 1-year and 4-year intervals. The model was trained on 31 multimodal features, including demographics, baseline KOOS scores, and radiographic measurements such as joint space width, osteophyte presence, and sclerosis grades. By generating personalized risk projections over clinically meaningful timeframes, this approach may help address an existing gap in KOA management, which is especially valuable given the heterogeneous progression of the disease. The model captures the known discordance between subjective symptoms and objective structural changes in KOA[19-21]. Jointly modeling function and structure enables improved risk stratification, facilitating the development of adaptive therapeutic strategies. The system demonstrates generalization across different time horizons, suggesting a degree of temporal stability in the evaluated tasks.

While recent research explores molecular biomarkers, genomics, and metabolomics for predicting KOA progression[10], these approaches face significant implementation barriers in clinical settings. Such methods often require invasive procedures, costly assays, or surgical specimens[24]. Furthermore, their availability in routine clinical environments remains limited, and center-specific biases and demographic variations frequently compromise their generalizability[22,25].

**Therapy Agents Group via Multi-Agent Collaboration**

The third core component of the KOM system utilizes a multi-agent architecture that generates personalized intervention plans tailored to individual patient clinical presentations and etiologies. This framework integrates specialized agents focused on distinct therapeutic domains, including exercise prescription, pharmacological and surgical interventions, nutritional planning, and psychological support. These domain-specific agents operate independently while being coordinated by a central clinical agent that synthesizes their outputs into a cohesive, individualized treatment strategy.

This multi-agent approach represents a departure from monolithic language models toward a modular design that enhances transparency, domain expertise, and decision quality. Comparative evaluations against leading language models demonstrated that KOM delivered superior performance across key clinical metrics, including recommendation accuracy, personalization, and actionability. The performance differential was most significant in complex cases involving multiple comorbidities or atypical presentations, where general-purpose models typically produced either overly generic or inconsistent recommendations. Our investigation of RAG with prompt engineering for incorporating external medical literature yielded limited performance improvements. In several evaluation categories, the RAG-enhanced model underperformed compared to its base language model, particularly in terms of clinical adaptability and semantic coherence. This limitation likely results from contextual inconsistencies caused by semantic drift in retrieved documents, a recognized challenge in current RAG implementations for specialized clinical reasoning tasks [26-28]. In contrast, the KOM multi-agent framework demonstrated effective capabilities in problem decomposition, domain-specific reasoning, and iterative refinement. These strengths align with findings from other domains: decentralized multi-agent reinforcement learning has shown improved task execution and generalization in complex network systems[29]. At the same time, Meta AI's Cicero system achieved expert-level performance in strategic gameplay through coordinated decision-making among specialized agents[30].

**Enhancing Clinical Process through AI-Clinician Collaboration**

Beyond its core capabilities, we evaluated the KOM system in collaborative scenarios with physicians to simulate real-world clinical workflows. Physicians used KOM for image interpretation and treatment planning assistance, with comparative analyses revealing that these AI-physician partnerships performed better within our evaluation settings than either component used independently across diagnostic accuracy, workflow efficiency, and treatment quality metrics. This collaborative approach exemplifies the human-AI symbiosis paradigm in healthcare, where AI systems function as intelligent assistants that enhance decision quality while reducing cognitive load.

Recent research strongly supports this collaborative model. Goh et al. [31] demonstrated that emergency physicians using an LLM achieved 15% higher diagnostic accuracy and greater decision consistency compared to unaided controls. In a multicenter study, clinicians using GPT-4 for patient case evaluation reported 20% faster processing times and a 12% improvement in review consistency. Similarly, Ayers et al. [32] demonstrated that a fine-tuned chatbot matched primary care physicians in terms of completeness and empathy when addressing common health questions. These findings underscore the practical value of AI-human collaboration, particularly in educational settings where structured guidance is crucial. By enabling physicians to explore complex reasoning with AI support, KOM functioned as a supportive tool for decision making in this study that helps develop diagnostic reasoning and decision-making skills.

We anticipate that AI-assisted clinical education will become fundamental to modern medical training. As language models improve in contextual understanding and interfaces become more intuitive, systems like KOM will evolve into intelligent learning partners, potentially transforming how clinical reasoning is taught and assessed[33-35].

**Limitation**

Despite these strengths, several limitations exist. First, while Assessment Agent performed well in retrospective testing, clinical implementation requires prospective validation with doctoral trainee oversight. Second, KOM focuses solely on knee osteoarthritis and may generate inappropriate outputs for other knee conditions; future versions should include a preliminary classifier to distinguish KOA from non-KOA pathologies. Third, the progression prediction model could benefit from additional factors beyond the current 31 features, which themselves present data collection challenges; wearable integration and feature optimization could address these issues. Finally, the intervention component lacks advanced mechanisms for resolving conflicting recommendations across different therapeutic domains, currently relying on basic prioritization strategies rather than sophisticated conflict resolution.

**Implication for Future Research**

In future research, we aim to enhance our agents while maintaining their clinical utility and effectiveness. Although our current integration of deep learning with large language models effectively addresses clinical needs, we seek to develop more compact models with higher integration capacity that could potentially function on mobile devices for real-time assessment. Regarding diagnostic evaluation, we plan to implement embedded technologies for disease monitoring, which could reduce dependence on traditional radiographic examinations that require specialized equipment, thereby streamlining the assessment process. For the prediction component, we plan to conduct further parameter analysis to identify more readily obtainable and clinically relevant variables. Regarding treatment recommendations, we intend to validate the efficacy of AI-generated management plans compared to conventional approaches through controlled clinical studies. Building upon our KOA management framework, this methodological approach can be adapted to other chronic conditions that require comprehensive management strategies.

**Conclusion**

This study presents KOM, a multi-agent artificial intelligence system for precision management of knee osteoarthritis. It successfully performs patient information collection, disease assessment, progression prediction, risk factor identification, and generates management plans. In our evaluation, it performed better than the general-purpose AI models tested on the specified setting tasks. The three-arm comparative study demonstrated that doctoral trainee-KOM collaboration achieved superior performance while reducing processing time. This study establishes a methodological foundation for developing scalable, evidence-based management strategies that may be adapted to address other chronic disorders.

## Methods

### Ethical Considerations and Study Design

For this study, we retrospectively recruited 300 patients with KOA from West China Hospital, Sichuan University. Among them, 250 cases with complete baseline and follow-up data were included for retrospective validation of the treatment planning module. In addition, a subset of 50 de-identified cases was selected to construct a simulated cohort for prospective evaluation under controlled experimental conditions. The Ethics Committee of West China Hospital approved the study protocol (approval number: 23-2277). Radiographic data for deep learning model development were obtained exclusively from the publicly accessible OAI database [36], a NIH-funded longitudinal observational study that provides standardized bilateral knee radiographs with corresponding clinical and demographic data; the OAI dataset is publicly licensed for research purposes and did not require additional approval.

All chatbot queries were conducted between January and April 2025 in Chengdu, Sichuan, China. For performance evaluation, three senior orthopedic and sports medicine experts independently assessed model outputs under a double-masked design in which evaluators were unaware of model identities; no patients or public participants were involved. In addition to accuracy and consistency, we screened model outputs for potentially harmful, misleading, or biased responses (eg, unsafe medication recommendations, diagnostic errors, or demographic bias) and found none. Apart from the OAI data, which remains under its original license, all other datasets and code used in this study are owned by the research team; however, our example code and study prompts have been made publicly available to promote transparency and reproducibility. In the development of KOM, we initial attempted using single-agent to perform all tasks in the KOA care pathway. But it performed poorly, which leads to the adoption of a multi-agent strategy in which each agent was specifically trained for different tasks. Through iterative rounds of testing and development the structure of KOM system was finalized, which included the Assessment Agent, Risk Agent, and Therapy Agents Group.

**Assessment Agent**

Functionality and Workflow

Upon accessing the KOM system, patients first indicate whether bilateral knee anteroposterior radiographs are available. When provided, the system automatically performs deep learning-based analysis of each knee, generating assessments of osteoarthritis severity, joint space narrowing, subchondral bone sclerosis, and osteophyte presence.

Subsequently, the intelligent conversational interface collects structured patient information, encompassing demographics, chief complaints, medical and family histories, current treatments, and lifestyle factors such as physical activity, occupational loading, and prior joint injuries. It also gathers data on metabolic and hormonal status, psychological and nutritional health, and treatment preferences. The interface identifies and prompts for missing information to ensure data collection. Patients can request clarification on medical terminology or data requirements in real-time. For unavailable clinical assessments such as the KOOS score, the Assessment Agent guides patients through the complete assessment protocol.

Development of the image analysis module

The imaging analysis models were trained on 12,719 bilateral knee radiographs obtained from the publicly available OAI database. The pipeline first performs joint localization using two independently trained U-Net models, which define the regions of interest for subsequent processing. Multi-task image analysis is then conducted using eleven task-specific deep neural networks based on the ResNet architecture, enabling simultaneous classification of KOA severity, joint space narrowing, subchondral sclerosis, and osteophyte presence.

Knee Joint Localization with an Enhanced UNet Pipeline

To automatically identify the central regions of bilateral knees in radiographs, we implemented an optimized UNet[37]-based segmentation framework designed explicitly for high-contrast X-ray images. This localization process served as the foundation for all subsequent image analyses.

We trained the model on 200 manually annotated anteroposterior radiographs, each paired with binary masks outlining knee centers. The architecture featured a five-layer UNet with skip connections, batch normalization, and ReLU activations. We initialized weights using Kaiming normalization to promote stable convergence. To address the significant foreground-background imbalance typical in knee radiographs, we employed an equally weighted hybrid loss function combining binary cross-entropy and Dice loss. During training, input images and masks underwent synchronized random flipping and resized cropping to maintain alignment. Performance evaluation utilized two metrics: bounding box intersection-over-union (IoU) and center point localization error. The Hungarian algorithm matched predicted and ground truth regions, ensuring unbiased metric computation. Training proceeded for 40 epochs using the Adam optimizer with cosine annealing learning rate scheduling, with model selection based on validation IoU performance.

Radiographic Classification of KOA Features Using Transfer-Learned ResNet50

To facilitate multidimensional classification of KOA-related radiographic features, we implemented a ResNet[38]-50-based deep learning pipeline trained on localized anteroposterior (AP) knee radiographs. Following initial joint localization, each cropped image was processed through a modified ResNet-50 architecture, where the final classification layer was replaced with task-specific output heads.

Classification Targets and Dataset Construction

We curated a series of supervised classification tasks covering 11 clinically relevant radiographic features: KOA severity; Joint space narrowing of the medial and lateral compartments; Subchondral bone sclerosis (medial femur, lateral femur, medial tibia, lateral tibia); Osteophyte presence (binary detection across the same four compartments). For each task, we constructed balanced datasets containing 500 images per severity level. The data was partitioned into training, validation, and testing sets using an 8:1:1 ratio to ensure a methodical approach to model development and evaluation.

KL Grading Adjustment and Label Consolidation

While the Kellgren-Lawrence (KL) grading system represents the established standard for radiographic KOA severity assessment[39], model development revealed consistent difficulties in discriminating between KL grades 0 and 1. Expert review, conducted by two musculoskeletal radiologists and a senior orthopedic surgeon, confirmed that the visual differences between these grades were subtle and had minimal implications for treatment[40,41]. Consequently, we consolidated KL 0 (normal) and KL 1 (doubtful) into a unified None/Doubt category. The remaining KL grades were mapped as follows: KL 2 → Mild, KL 3 → Moderate, KL 4 → Severe. This restructuring produced a clinically meaningful and computationally effective four-class KOA severity framework:

None/Doubt (KL 0–1)

Mild (KL 2)

Moderate (KL 3)

Severe (KL 4)

To maintain compatibility with clinical standards, multimodal models were initially trained on the original five-grade KL classification and subsequently converted to the four-grade schema during downstream evaluation.

All radiographs underwent standardized bone window processing (center: 300 HU, width: 1500 HU) to enhance bony structure contrast and minimize irrelevant soft-tissue variation. Images were normalized to the [0, 1] range, resized to 256 × 256 pixels, and randomly cropped to 224 × 224 pixels during training. Data augmentation included horizontal flipping, small-angle rotations, and brightness/contrast adjustments, applied synchronously to maintain label consistency. Model training was performed using a five-fold cross-validation approach. For each fold, we optimized the network using the Adam optimizer (initial learning rate: 1e-5) with a stepwise decay schedule ($\gamma = 0.1$ every seven epochs). Cross-entropy loss was used as the objective function. An early stopping mechanism (patience: 10 epochs; minimum delta: 1e-6) preserved the best-performing model state based on validation loss.

Ground truth annotations from the OAI dataset were subjected to multi-level expert validation to ensure labeling accuracy and inter-rater consensus. To mitigate overfitting and maximize generalization, only center-cropped images were used during validation and testing.

Model performance was evaluated using classification accuracy, confusion matrices, and area under the receiver operating characteristic curve (AUC-ROC) across all classes. These metrics were calculated per fold and averaged to obtain robust task-specific performance estimates. To enhance interpretability, Gradient-weighted Class Activation Mapping (Grad-CAM) was applied to visualize spatial attention maps on representative validation samples. These visualizations confirmed the appropriate model focus and identified anatomical regions that informed specific predictions. Training histories, including per-epoch loss, accuracy curves, confusion matrices, and ROC plots, were documented to ensure transparency and reproducibility.

Development of KOA Patient Information Collection module

To facilitate standardized and clinically interpretable documentation of KOA patient profiles, we developed an Assessment Agent based on a LLM optimized through domain-specific prompt engineering. The agent was designed to function as a clinical intake assistant, conducting dynamic natural language dialogue to compile clinical profiles for each patient.

System Design and Implementation

The generation process began with a structured system prompt that established the documentation template and directed the model to gather information across 11 essential clinical domains: demographics, chief complaint and history of present illness, radiographic findings, past and family history, current treatment status, psychological well-being, nutritional condition, treatment goals and preferences, and available rehabilitation resources. To enhance clinical utility, the prompt incorporated mechanisms for real-time clarification of medical terminology. When simulated patients expressed uncertainty regarding specialized concepts (e.g., KOOS scoring or subchondral sclerosis), the LLM automatically provided context-appropriate explanations before continuing the assessment dialogue.

Evaluation Methodology

We evaluated the Assessment Agent using 100 simulated KOA patient scenarios from West China Hospital. For each simulation, a trained research assistant assumed a predefined patient persona and engaged in an interactive dialogue with the LLM. Each session proceeded until the model determined that sufficient clinical information had been collected, at which point it autonomously concluded the interview process. All generated outputs were anonymized and randomized for subsequent expert review.

To assess the clinical quality of the generated structured cases, we conducted a blinded expert evaluation. Three senior orthopedic and sports medicine physicians independently evaluated each case across four predetermined dimensions:

Field completeness: Assessment of whether all expected clinical fields contained adequate information

Logical consistency: Evaluation of internal coherence and clinical plausibility of the narrative

Medical accuracy: Assessment of appropriate and correct application of clinical terminology and judgment

Readability: Evaluation of clarity, fluency, and professional expression

Before formal assessment, all expert evaluators participated in a calibration session.

This session included review and discussion of representative examples illustrating high-, medium-, and low-quality cases, followed by collaborative refinement of the scoring rubric. During the formal evaluation phase, each expert was blinded to both the origin and sequence of the cases, and no communication was permitted during individual assessments. Each dimension was rated on a 1-5 scale, with the final score for each case calculated by averaging ratings across the three evaluators.

Agent Benchmarking and Comparative Evaluation

To assess the diagnostic performance of the KOM system relative to current vision-language models, we conducted a benchmarking study comparing five leading multimodal large language models (MLLMs): Google Gemini 2.0 Pro, GPT-4o, Claude 3.7 Sonnet, QwenMax VL, and LLaMA 3.2 90B Vision Instruct. All models were evaluated under identical input and task conditions to ensure methodological consistency.

Evaluation Dataset and Protocol

The evaluation dataset consisted of 500 bilateral knee radiographs (1,000 knees in total) obtained from the publicly available OAI cohort. All samples were excluded from any previous training or fine-tuning of the KOM system or its constituent components. For each case, a standardized input prompt was constructed to elicit three clinically relevant outputs:

Bilateral KOA severity grading (based on the Kellgren-Lawrence scale)

OA presence detection for each knee (binary classification)

Left knee localization (spatial discrimination to assess model orientation awareness)

To maintain consistency across model evaluations, all knee radiographs underwent preprocessing to ensure uniform viewing orientation (with the left knee positioned on the right side, facing the observer), and standardized diagnostic prompts were employed. Each model received identical image-text inputs and was evaluated based solely on its unmodified output without manual intervention.

Evaluation Metrics and Ground Truth

For KOA severity grading, the reference standard was derived from the revised OAI dataset and mapped into a four-class severity framework: None/Doubt (KL 0-1), Mild (KL 2), Moderate (KL 3), and Severe (KL 4). The KOM system was explicitly designed to predict KOA severity grades that align directly with this classification schema. All other multimodal agents were first prompted to produce KL grades, which were then converted into corresponding severity categories using the same KL-to-severity mapping protocol.

For the OA presence detection task, models employed different output approaches. The KOM system internally predicted a KL severity grade for each knee, from which OA presence was derived using a predefined threshold; cases classified as KL 0 or 1 were categorized as "No OA." At the same time, KL $\geq 2$ was designated as "OA present." In contrast, multimodal LLMs were prompted to directly determine whether OA was present or absent for each knee, without generating an explicit KL grade.

To enable valid comparison, ground truth OA labels were derived using consistent KL-based thresholding: KL 0-1 $\rightarrow$ No OA, KL 2-4 $\rightarrow$ OA present. Model predictions were binarized accordingly, and accuracy was calculated separately for left and right knees.

**Risk Agent**

Functionality and Workflow

Upon activation of the symptom and radiographic prediction agent, patients provide baseline data encompassing 31 clinical parameters, including body mass index (BMI), age, body weight, KOOS subscale scores, and bilateral knee muscle strength measurements. These parameters can be seamlessly populated from the previously completed structured clinical documentation Agent or directly entered by the patient.

Symptom Prediction Submodule

To predict KOOS subscale outcomes at 1-year and 4-year follow-ups, regression models utilizing XGBoost[42], LightGBM[43], Random Forest[44], Gradient Boosting[45], Support Vector Regression (SVR[46]), and Elastic Net[47] algorithms were developed.

The input dataset comprised 31 patient parameters that underwent preprocessing, including removal of incomplete cases, categorical encoding, and z-score standardization. Model performance was evaluated using five-fold cross-validation, which employed multiple metrics: $R^2$, mean squared error (MSE), mean absolute error (MAE), and Pearson correlation coefficient (Pearson r). Feature importance analyses were conducted and visualized via bar plots, residual analyses, and scatter plots to provide insights into model predictions. Quantitative evaluation results are presented in the supplementary Graphs S3.

Radiographic Prediction Submodule

For radiographic outcome prediction, structured clinical data from the OAI dataset were utilized to forecast KL grades for both knees at 1-year and 4-year intervals, constituting four distinct prediction tasks. The dataset underwent stratified splitting (70% training, 30% validation) and class balancing to ensure uniform representation with 1,000 cases per KL grade. Eight machine learning algorithms were evaluated:XGBoost[42], LightGBM[43], Random Forest[44], Gradient Boosting[45], AdaBoost[48], Support Vector Machine (SVM[49]), K-Nearest Neighbors (KNN[50]), and Multi-layer Perceptron (MLP). Model robustness was assessed using 100 iterations of Monte Carlo cross-validation, with performance quantified through accuracy (ACC), weighted precision, recall, F1-score, and macro-area under the receiver operating characteristic curve (macro-AUC). Confusion matrices and ROC curves illustrating misclassification patterns and model discriminative capabilities are provided in the Supplementary Graphs S3.

Risk Factor Analysis Submodule

To produce individualized risk factor assessments, predictive models were enhanced with SHAP[51] analyses. These analyses produce interactive force plots that illustrate the relative contributions of key clinical parameters, such as BMI, body weight, age, and muscle strength, in predicting patient-specific osteoarthritis progression risk.

Each prediction task utilizes the single best-performing model, allowing comparative analyses of risk factor contributions. This flexible analytical framework supports both consensus-driven and divergent risk assessments, providing clinicians with insights to tailor interventions to individual patient profiles.

**Therapy Agents Group**

Functionality and Workflow

When entering the multidisciplinary Intervention Agent, patient profiles that integrate structured clinical data from the Assessment Agent and personalized predictive outcomes with risk factor analyses produced by the Risk Agent are used as foundational inputs. This Agent employs a collaborative multi-agent artificial intelligence architecture to generate tailored therapeutic recommendations across diverse clinical domains autonomously. The system incorporates specialized intelligent agents functioning as exercise prescriptionists, surgical and pharmacological interventionists, and nutritional and psychological specialists. Each agent independently develops structured, evidence-informed treatment recommendations within its domain of expertise. A clinical decision-making agent subsequently synthesizes these domain-specific interventions to produce a patient-specific management strategy. Throughout this process, the system prioritizes clinical applicability, scientific accuracy, patient safety, and individualized care. Detailed metrics for evaluating the clinical effectiveness of these interventions are described in subsequent sections.

Knowledge Base Construction

To ensure evidence-based therapeutic recommendations, extensive domain-specific knowledge bases were developed, encompassing five key intervention categories:

exercise rehabilitation, surgical techniques, rehabilitation interventions, nutrition, and psychological therapies. A literature search conducted in the PubMed database initially identified 33,641 articles and clinical guidelines. Through article evaluation, these were refined to a core repository of 4,017 high-quality, peer-reviewed publications and internationally recognized clinical guidelines. Each selected document underwent targeted extraction of results and recommendations sections, excluding unrelated content to optimize relevance and retrieval accuracy. The annotated excerpts were organized into structured repositories optimized for RAG, thereby enhancing the accuracy and specificity of knowledge retrieval and agent-generated therapeutic recommendations. Complete details of literature selection criteria, evaluation methodologies, and knowledge base composition are provided within the supplementary materials.

Development of Individual and Multi-Agent Architectures

The multidisciplinary clinical recommendation system integrates four distinct intelligent agents, each leveraging the Qwen-Max large language model as a foundational engine, further optimized through targeted RAG techniques and tailored prompt engineering.

Exercise Prescriptionist Agent: applies the FITT-VP framework (Frequency, Intensity, Time, Type, Volume, and Progression), dynamically customizing exercise regimens based on patient-specific clinical data and therapeutic objectives retrieved via RAG.

Surgical and Medication Specialist Agent: Matches patient profiles against surgical guidelines and pharmacological guideline databases, providing detailed and precise intervention proposals with appropriate dosing, timing, and procedural specifications.

Nutritional and Psychological Specialist Agent: Integrates nutritional prescriptions strictly adhering to the ABCMV principles (Adequacy, Balance, Calorie control, Moderation, Variety), supplemented by tailored psychological management strategies responsive to individual patient needs.

Clinical Decision-Making Agent: Serves as the integration hub where outputs from the specialized agents converge. This agent critically evaluates and synthesizes the recommendations, optimizing outcomes along dimensions of accuracy, comprehensiveness, personalization, patient safety, and domain-specific professional standards.

Through this architecture, the final integrated prescription is validated and tailored to each patient's unique clinical profile and therapeutic requirements.

**Clinical Validation with Real-World Patient Data**

We retrospectively assembled a cohort of 250 KOA patients from West China Hospital, Sichuan University. This dataset captured detailed baseline demographics, clinical examinations, radiographic findings, etiological diagnoses, patient treatment preferences, and institutional resource availability. Clinical management strategies were classified into five categories:

Conservative management (n = 73)

Total knee arthroplasty (n = 62)

Unicompartmental knee arthroplasty (n = 43)

Osteotomy (n = 39)

Arthroscopic surgery (n = 33)

Evaluation Protocol

For each patient case, the KOM system, GPT-4o, Claude 3.7, and five additional vision-language models independently generated treatment recommendations. To eliminate bias, all model outputs were de-identified and randomized into a single pool for evaluation. Three senior orthopedic experts, each with over ten years of specialized clinical experience, conducted fully blinded evaluations of all recommendations using a unified seven-dimensional rubric.

Evidence-based practice

Completeness

Exercise prescription

Nutrition prescription

Personalization

Accessibility and feasibility

Safety

Before formal scoring, the reviewers participated in a calibration workshop where they jointly reviewed five exemplar cases representing both exemplary and suboptimal recommendations. This process resolved scoring discrepancies and resulted in a detailed evaluation manual, ensuring harmonized threshold definitions and high inter-rater consistency. In parallel, standard clinical prescriptions were developed as benchmark references, with comparative linguistic similarity analyses (including BLEU, BERT, and ROUGE metrics) quantifying each model's adherence to the gold-standard clinical treatment protocol.

Prospective Evaluation Using a Simulated Patient Cohort

Study Design and Participant Selection

We extracted de-identified records for 50 KOA patients from the West China Hospital database. Baseline weight-bearing radiographs and accompanying clinical data underwent independent review and confirmation by two senior orthopedic surgeons; only cases approved by both reviewers advanced to the simulated evaluation. Twenty doctoral candidates without prior KOA-specific imaging or therapeutic training (age 25–35 years, ≤1 year clinical rotation) were randomized via a computerized draw application into two arms (n = 10 each):

"Physicians-only" group

"Physicians + KOM" group

Randomization was performed by an independent data manager using sealed electronic envelopes to ensure allocation concealment. Before case evaluation, all participants attended a single standardized training session that covered KOA radiographic interpretation and prescription formulation protocols.

Clinical Materials and Assessment Methodology

Corresponding radiographic data for each simulated case were drawn from the OAI, ensuring robust clinical validity and standardized imaging support. Radiographic grading accuracy was calculated as the proportion of cases for which the predicted KOA severity exactly matched the adjudicated reference grade in our corrected OAI database. We recorded the total time required for radiographic interpretation and prescription development tasks across all groups, quantitatively assessing efficiency gains attributable to KOM system integration.

Treatment Plan Evaluation

To evaluate clinical decision-making quality, we generated treatment plans for 50 simulated KOA patient cases across three cohorts:

KOM group (KOM runs three times per case)

physicians' group (three different physicians per case)

collaboration group (three physicians using KOM)

This process resulted in a total of 450 de-identified plans. All plans were pooled, randomized, and stripped of origin labels to prevent evaluation bias.

Two senior orthopedic specialists, each with over a decade of specialized experience, independently scored every plan using a harmonized seven-dimensional rubric:

Evidence-based practice

Completeness

Exercise prescription

Nutrition prescription

Personalization

Accessibility and feasibility

Safety

Before formal review, the experts jointly examined five exemplar plans (representing both high and low quality), reconciled scoring discrepancies, and finalized a detailed evaluation manual to ensure consistent application of rating thresholds.

To anchor assessments in best-practice care, a third senior specialist created gold-standard prescriptions for each case based on current clinical guidelines. Finally, we quantitatively compared each free-text plan against its benchmark using established textual similarity metrics (BLEU, BERT, and ROUGE), enabling a appraisal of each plan's coherence with expert protocols.

procedural steps, the full scoring rubric, and calibration details are provided in the Supplementary Methods.

**Statistical Analysis**

To standardize scores within each model across metrics, we applied row-wise z-score normalization to the model-by-metric mean matrix used for visualization (Fig. 4f). For model m and metric d, with mean score $S_{m,d}$, we computed

$$z_{m,d} = \frac{S_{m,d} - \mu_m}{\sigma_m} \quad (1)$$

, where $\mu_m$ and $\sigma_m$ denote the mean and standard deviation of all metric scores for model m, respectively. This emphasizing the relative distribution of metrics within a model rather than between-model differences. The resulting z-score matrix was used for heatmap visualization and pattern analysis.

For statistical analysis of between-group differences, diagnostic accuracy and task completion time were treated as continuous variables and assessed for normality using the Shapiro-Wilk test. For normally distributed data, independent samples t-tests were used; for non-normally distributed data, the Mann-Whitney U test was employed for between-group comparisons. Treatment quality scores, being ordinal variables, were consistently analyzed using Mann-Whitney U tests. Statistical significance for all between-group comparisons was established at $p < 0.05$.

**Acknowledgements**

**Funding**

This work was funded by the Youth Research Fund of Sichuan Science and Technology Planning Department. Grant number 23NSFSC4894 (received by Xi Chen)

**Author Contributions**

W.L., X.C., and Z.J. are the main designers of the study. W.L., X.C. are the main executors of the study. K.L., and J.L. contributed to the study by managing and supervising the revision work and providing critical feedback during the major revision process. H.Z., H.C., K.L., and W.L. served as consultants for computer science-related knowledge. H.C. and W.L. developed the code for this study and performed the model training. W.L., X.C., Z.J., L.Z., K.Z., R.T., L.W., and M.Y. evaluated the models' responses and prepared the test dataset. W.L., X.C., and Z.J. participated in drafting the manuscript. K.L., and J.L. provided overall guidance and supervision for the project. All authors have read and approved the final version of the manuscript.

**Competing interests**

The authors declare no competing interests.

**Ethics declarations**

This study was approved by the Ethics Committee of West China Hospital, Sichuan University (Approval No. 23-2277). All procedures complied with the Declaration of Helsinki and relevant national regulations, including China's Personal Information Protection Law.

**Code availability**

The complete source code for this project is publicly accessible at https://github.com/jacobliuweizhi/KOM. A demonstration of the implementation is available through our interactive web interface at https://huggingface.co/spaces/Miemie123/Streamlit?page=Tailored+Therapy+Recommendation&start=1.

**Data availability**

The code developed for this study is available at: https://github.com/jacobliuweizhi/KOM under the GNU Affero General Public License v3.0. W.L., H.Z., H.C., and K.L. contributed to the code development and are responsible for maintaining the repository. Reference documents used in the RAG module are listed in the repository. Osteoarthritis-related imaging and clinical data used in this study are accessible through the OAIdatabase (https://nda.nih.gov/oai), subject to data use agreements.

**Figure legend**

Figure 1.

Title: Multi-Agent Framework for KOA Assessment, Risk Prediction, and Therapy Planning

Caption: Overview of the multi-agent framework for KOA management, clinical workflow, and performance benchmarking. The framework includes an Assessment Agent, Risk Agent, and Therapy Agents Group working together to evaluate symptoms and imaging, predict progression, and plan personalized therapy. The workflow involves assessment, risk prediction, and therapy planning phases. KOM showed higher performance than the evaluated baseline models and participants within the specific tasks and datasets assessed in this study.

Figure 2.

Title: Performance Evaluation and Benchmarking of the KOA Assessment Agent.

Caption: KOM outperforms major large models in KOA severity grading and OA detection across severity levels. Training and visualization demonstrate accurate joint localization and radiographic focus. KOM achieves better performance on multiple assessment tasks and high expert evaluation scores for information quality, with superior accuracy for both knees compared to competitors.

Figure 3.

Title: Performance Evaluation and Predictive Modeling of the KOA Risk Agent.
Caption: The framework integrates structured data for KOOS and KL grade prediction. Various machine learning models are compared using accuracy, confusion matrices, and radar plots across time points. Performance varies across models, with some showing stronger long-term prediction capabilities.

Figure 4.

Title: Performance Evaluation and Comparative Analysis of the KOA Therapy Agents Group

Caption: KOM achieves the highest expert scores and rankings across seven clinical domains compared to multiple large models. Text similarity analysis shows KOM outputs most closely match reference prescriptions across BLEU, ROUGE, and BERT metrics.

Figure 5.

Title: Evaluation of Human–AI Collaboration in KOA Diagnosis and Treatment

Planning

Caption: Study design, physician–KOM collaboration workflow, diagnostic efficiency, clinical evaluation, and text similarity analysis across different groups. The study included 50 KOA patients (KL 0–4, n = 10 each) and compared three settings: physician-only (MS), KOM-only, and physician–KOM collaboration (MS+KOM). Panel (a) shows the study design; (b) illustrates the physician–KOM collaboration process; (c) presents diagnostic time comparisons, with the collaboration group achieving faster completion; (d) shows expert evaluations across seven clinical criteria, where the collaboration group achieved the highest scores; (e) compares grading accuracy, which improved from MS to KOM and was highest in MS+KOM; and (f) presents text similarity metrics (BLEU, ROUGE, BERT), showing that collaboration produced outputs closest to reference prescriptions.